\documentclass{article}

\usepackage{setspace} 

\usepackage{PRIMEarxiv}

\usepackage[utf8]{inputenc} 
\usepackage[T1]{fontenc}    
\usepackage{hyperref}       
\usepackage{url}            
\usepackage{booktabs}       
\usepackage{amsfonts}       
\usepackage{nicefrac}       
\usepackage{microtype}      
\usepackage{lipsum}
\usepackage{fancyhdr}       
\usepackage{graphicx}       
\usepackage{amsmath}
\usepackage[table,xcdraw]{xcolor}
\usepackage{booktabs}
\usepackage{graphicx}
\graphicspath{{media/}}     

\title{Cognitive network science quantifies feelings expressed in suicide letters and Reddit mental health communities
}

\author{
  Simmi Marina Joseph \\
  CogNosco Lab, Department of Computer Science\\
  University of Exeter\\
  Exeter, UK\\
  \texttt{ sj504@exeter.ac.uk } \\
   \And
  Salvatore Citraro\\
  Department of Computer Science, University of Pisa \\
  Largo Bruno Pontecorvo, 3, Pisa \\
  KDD-Lab (ISTI-CNR)\\
  G. Moruzzi, 1, Pisa \\
  \texttt{salvatore.citraro@phd.unipi.it}
  \And
  Virginia Morini \\
  Department of Computer Science, University of Pisa\\
  Largo Bruno Pontecorvo, 3, Pisa \\
  KDD-Lab (ISTI-CNR)\\
  G. Moruzzi, 1, Pisa \\
  \texttt{virginia.morini@phd.unipi.it} 
  \And
  Giulio Rossetti \\
  KDD-Lab (ISTI-CNR)\\
  G. Moruzzi, 1, Pisa \\
  \texttt{giulio.rossetti@isti.cnr.it} 
  \And
  Massimo Stella \\
  CogNosco Lab, Department of Computer Science\\
  University of Exeter\\
  Exeter, UK\\
  \texttt{ m.stella@exeter.ac.uk } \\
}

\begin{document}
\maketitle

\begin{abstract}
Writing messages is key to expressing feelings. This study adopts cognitive network science to reconstruct how individuals report their feelings in clinical narratives like suicide notes or mental health posts. 
We achieve this by reconstructing syntactic/semantic associations between concepts in texts as co-occurrences enriched with affective data. 
We transform 142 suicide notes and 77,000 Reddit posts from the r/anxiety, r/depression, r/schizophrenia, and r/do-it-your-own (r/DIY) forums into 5 cognitive networks, each one expressing meanings and emotions as reported by authors. 
These networks reconstruct the semantic frames surrounding "feel", stem for "to feel" and "feelings", enabling a quantification of prominent associations and emotions focused around feelings. 
We find strong feelings of sadness across all clinical Reddit boards, added to fear r/depression, and replaced by joy/anticipation in r/DIY.
Semantic communities and topic modelling both highlight key narrative topics of "regret", "unhealthy lifestyle" and "low mental well-being". 
Importantly, negative associations and emotions co-existed with trustful/positive language, focused on “getting better”. 
This emotional polarisation provides quantitative evidence that online clinical boards possess a complex structure, where users mix both positive and negative outlooks. 
This dichotomy is absent in the r/DIY reference board and in suicide notes, where negative emotional associations about regret and pain persist but are overwhelmed by positive jargon addressing loved ones. 
Our quantitative comparisons provide strong evidence that suicide notes encapsulate different ways of expressing feelings compared to online Reddit boards, the latter acting more like personal diaries and relief valve. 
Our findings provide an interpretable, quantitative aid for supporting psychological inquiries of human feelings in digital and clinical settings.
\end{abstract}


\doublespacing

\section{Introduction}
Humans' cognition has no equal. 
Our structured emotions and keen social attitudes lead us to develop sympathy, altruism and cooperation, among the wide range of social ties we can form \cite{dunbar1993coevolution}. 
However, there exists a downside of such a natural social willingness. 
It concerns individuals’ difficulties to get integrated into the shared values of a culture, an anomie \cite{durkheim2020suicide} or anomaly that also concerns individuals’ inability to understand, express or regulate their own emotions. 
Complex circumstances can lead to unbearable situations and towards behaviours that are unique in nature, where people could reinforce negative emotions and isolate themselves \cite{tugade2004psychological}. 
The most extreme consequences of such behaviours lead to drastic, often non-reversible, solutions \cite{batinic2017comorbidity}. Among them, the willingness to die is the most alarming and tragic one.
According to the World Health Organization, more than 7,00,000 people commit suicide every year, and it is reported as the fourth leading cause of death among 15–29-year-olds\footnote{“Mental Health and Substance Use.” \url{https://www.who.int/teams/mental-health-and-substance-use/suicide-data} (accessed Mar. 22, 2021)}. 
Suicide can be experienced as the ultimate step of a long period of distress caused by clinical pathologies like anxiety or depression or by clinical psychosis like schizophrenia \cite{gaur2021characterization}. 
Victims' reluctance to speak about their feelings to a near person makes suicide hard to predict and understand \cite{pavalanathan2015identity}. 
Nevertheless, interestingly, people tend to be more inclined to use online social platforms to talk about their own deeply intimate emotions to other people who share similar problems \cite{de2014mental, de2016discovering}, a human tendency analogous to homophily in social networks analysis, i.e. a bias for social ties to be established between similar actors \cite{mcpherson2001birds}. 
Due to its forum structure as well as users’ anonymity, Reddit comes as one of the most promising social platforms to characterize mental health communities i.e., actual pools where people aim to gain information about their health problems, share their symptoms and difficulties, or give support to others having similar problems \cite{de2014mental, gkotsis2016language}. 
Several recent studies have started using clinical boards on Reddit as data sources for identifying suicide ideation and other disorders linked with emotional distress \cite{de2016discovering, yoo2019semantic, ji2018supervised, kim2020deep} and we proceed along this direction.

In this paper, we aim to understand both online users’ and suicide victims’ language to uncover their underlying emotional patterns and quantify their degree of similarity and variability. 
In the era of Big Data, healthcare systems are increasingly supported by Artificial Intelligence (AI) tools that can allow to analyze suicide note contents and extract patterns and features from texts \cite{ji2018supervised, tadesse2020detection, gaur2021characterization} but often at the cost of losing information about the interpretation of a machine learning classification/regression model \cite{rudin2019stop}. 
We provide a novel, interpretable approach to understand and compare the thoughts of suicide victims with those of individuals experiencing mental health conditions. In particular, we leverage a simple, interpretable, cognitive network methodology that directly compares individuals' feelings when expressing their narrative in suicide notes and clinical Reddit boards. 
This method may open new ways for clinicians to deeply understand the relatability of suicide with any mental condition, exploring further possibilities to move towards the direction of predicting suicidal tendencies in the vulnerable category. 
It can also open additional ways to detect if a suicide victim had any potential mental disorders.

The rest of the paper is organized as follows. 
In Section \ref{sec:background}, we provide a broad background framing this work. Section \ref{sec:methods} describes the methods used to exploit our analysis, spanning from novel cognitive network science to classic NLP tools, together with the datasets used for the analysis. 
Section \ref{sec:results} sums up the main results of the paper, while Section \ref{sec:discussion} discusses the most promising directions of our work as well as its limitations and future lines of research. 

\section{Related Works}
\label{sec:background}
Text represents an insightful fragment of the psychology of individuals \cite{murray2003narrative, stella2020text, jackson2020text}. 
Within the context of clinical text analysis, several works trying to analyze and understand the content of suicide notes focused on finding dominant emotional words  (e.g., “love”) or the reason behind the gesture \cite{brevard1990comparison, foster2003suicide, handelman2007content}. 
In most of the analyzed letters, individuals express the willingness to escape from the pain and the anger towards other people and, interestingly, letters from completed suicides express more self-blame than those from attempted suicides \cite{brevard1990comparison}. 
Early studies also tried to characterize the differences between the text features hidden in suicide notes compared to other types of text.
As an example, an early study on a corpus of 66 suicide notes where half were genuine and half were simulated aimed to identify hidden features that can classify genuine and fake notes \cite{shneidman1957some}. 
More recently, the study of mental health discussions and suicide contents in online pools focuses predominantly on Natural Language Processing (NLP) techniques, often aiming to extract linguistic features for classification, e.g., predicting whether a textual content contains early signs of suicidal ideation \cite{schoene2016automatic, ji2018supervised, tadesse2020detection} or more in general, signs of mental health disorders \cite{shen2017detecting, kim2020deep, low2020natural}. 
For instance, Nikfarjam and colleagues \cite{nikfarjam2012hybrid} defined a framework to label 900 genuine suicide notes with their underlying emotions by firstly extracting syntactic and semantic features, thus using them to define the input rules for a machine learning classifier. 
Similarly, Low and colleagues \cite{low2020natural} rely on regression and 90 text features derived from Reddit (e.g., sentiment analysis, semantic categories, personal pronouns) to verify whether the COVID-19 pandemic was impacting mental balance. 
It is relevant to notice that most of those works \cite{shen2017detecting, ji2018supervised, kim2020deep} leverage as primary source Reddit mental health communities since they occur as a helpful source for monitoring people’s perceptions about their conditions through the emotional language they use to convey feelings.\newline
However, a recent discussion has underlined the risk of dehumanizing individuals while studying mental health prediction on social media \cite{chancellor2019human}. 
Indeed, using binary classification, as in the previous examples, and thus reducing mental health status to two corresponding machine learning classes (e.g.,  “positive”/“negative” or “suffering”/“not suffering”) produces severe biases that flatten the complexity of mental illness. 

Another branch of NLP that has been fruitfully used both in suicide notes as well as in online forums analysis to capture topical patterns in the mental health discourse is Topic Modeling \cite{rehurek2010software}. 
Topic modeling techniques are able to detect the hidden topics - lists of words - that characterize a set of textual documents, and this can help researchers in finding the events that led to suicide ideation or the most debated issues that grieve on individuals affected by mental health disorders \cite{fraga2018online, yoo2019semantic}. 
For instance,  Xue and her team leverage a popular topic modeling technique, i.e., Latent Dirichlet Allocation (LDA), to detect individuals' key themes and emotional reactions during the early stage of COVID-19 \cite{xue2020public}. 
The main limitation of this approach is that the definition of “topic” is loose and with no clear cognitive counterpart, since it is merely based on statistical co-occurrences that: (i) crucially depend on the partitioning of text across documents, (ii) can change according to an external parameter, i.e. an arbitrary number of desired topics, (iii) lead to multiple topics sharing significant fractions of words. 
Despite the presence of heuristics that can partially resolve these issues, LDA should be considered as a technique to use in synergy with other cognitively-grounded methods like word count inquiry or network analysis.

A promising tool for identifying emotional and semantic features in texts are Cognitive Networks \cite{castro2020contributions, stella2021cognitive}. 
Leveraging graph theory concepts, cognitive networks allow the analysis of and on complex structures in the human mind apt at storing and processing knowledge. 
Cognitive networks of conceptual associations aim to provide results that are interpretable from a cognitive point of view, e.g., being used in lexical retrieval tasks, early language learning studies, and, in general, whenever a cognitive process can be applied on a complex network structure \cite{siew2019cognitive, castro2020contributions, valba2021analysis}.
Such methods are transparent and help to characterize, among others, the most central words in a network as well as the positive or negative emotions surrounding word neighborhoods \cite{stella2020text}. 
Another important aspect of cognitive networks is their natural tendency to form modular clusters or communities \cite{siew2019cognitive}, i.e., sets of well-connected nodes more similar to each other rather than the rest of the graph, which can be exploited, among others, to study users’ sentiment on social media \cite{wang2017sentiment}. 
Applying cognitive networks on corpora of suicide notes can be useful for further understanding and getting meaningful insights into such rich text data, e.g., analyzing how some concepts tend to cluster and influence the overall structure of the network \cite{teixeira2021revealing}. 
In fact, the associations attributed to a given concept in text represent its so-called semantic frame \cite{fillmore2010frames}. 
Semantic frame theory indicates that meaning and perceptions relative to a concept can be reconstructed by analysing its semantic frame \cite{carley1993coding, fillmore2010frames} and cognitive networks enable this reconstruction in a quantitative way, allowing for researchers to reconstruct the structure of semantic/syntactic associations in texts and to focus attention over specific ideas reported in such networked structure \cite{carley1993coding, stella2020text, stella2020lockdown}.

In the following work we rely on a combination of methods from cognitive network science and NLP to study how people express their feelings through mental health online boards and genuine suicide notes.
\section{Methods}
\label{sec:methods}
This section outlines the key resources and methodologies adopted in this study.
\subsection{Textual data: Genuine suicide notes and clinical Reddit posts}
This work used letters and posts as personal narratives providing a glimpse into the psychology of individuals \cite{jackson2020text}. Specifically, we used suicide notes and Reddit blogposts from schizophrenia, depression, anxiety and “Do It Your own” (DIY). 

The 142 suicide notes used here were curated by Schoene and Dethlefs \cite{schoene2016automatic} and investigated also in \cite{teixeira2021revealing}. 
These English letters were authored by people who successfully took their lives. 
On average, a suicide note included 120 words. No additional information (e.g. socio-economic context) was available in the dataset. 
All notes are anonymised to protect the individuals’ privacy, in adherence with the Declaration of Helsinki.

Building our study on feelings expressed in suicide notes, we searched for other datasets sharing analogous clinical features. We selected a collection of 77,000 blog posts mined from Reddit by Kim and colleagues \cite{kim2020deep}. 
These texts indicated online posts on subreddits like r/schizophrenia, r/depression, and r/anxiety and were used by \cite{kim2020deep} to implement deep learning binary classifiers detecting the presence of clinically negative emotional states. The r/DIY subreddit was included as a reference poll, including texts that were produced online but coming from individuals not engaged in mental health discussions. 
These posts were scraped using ParseHub\footnote{\url{https://www.parsehub.com/}, Last Accessed: 08/10/2021} and anonymised by removing user details and names. 
In this instance, we focused only on these forums, though our approach can be extended to include multiple subreddit or online forums.

\subsection{Text cleaning, network construction and centrality, affective enrichment}
With NLTK in Python, texts were tokenized and reduced to sequences of sentences, each one containing a sequence of words. Word ordering was preserved but punctuation, symbols, hyperlinks and numbers were discarded. 
Individual words were stemmed, thus reducing the occurrence of multiple word forms relative to the same concept/idea (e.g. “sadness” and “sad” both became the stem “sad”). 
For stemming we used Porter’s algorithm as implemented in NLTK. 
With NLTK, stopwords excluding meaning negations - like “no” and “not” - were removed from all word lists. 
We then drew co-occurrence links between adjacent words, as performed in previous approaches using co-occurrence networks for authorship identification \cite{akimushkin2017text} or semantic framing in suicide notes \cite{teixeira2021revealing}. Co-occurrence links are also known as bi-grams, as they indicate the occurrence of two lexical items together. From a network perspective, counting the frequency of bi-grams was useful for filtering infrequent co-occurrences. However, once thresholded, we treated co-occurrence links as undirected and unweighted. 

\begin{figure*}[h]
\centering
  {\includegraphics[scale=0.75]{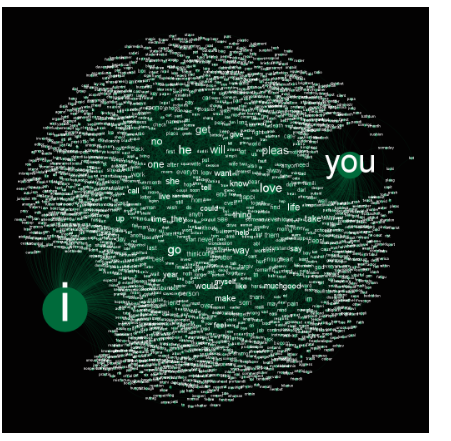}}
    \caption{Co-occurrence network of concepts as expressed in 139 genuine suicide notes.}
    \label{fig:fig_1}
\end{figure*}

Co-occurrence networks capture the syntactic structure linking concepts in sentences and attributing meaning to narratives \cite{carley1993coding, murray2003narrative}. However, when working with corpora differing in size and length (of words), it can be difficult to build networks possessing comparable/analogous connectivity among the same set of nodes. To overcome this limitation, we used suicide notes - our smallest dataset - as a baseline. It contained 6465 bi-grams/links. We then performed the same co-occurrence/bi-gram counts for other corpora and considered only the top-ranked 6465 bi-grams to build co-occurrence networks representing the narratives of such online clinical populations. This hard constraint enabled the construction of networks possessing equivalent link connectivity and it filtered strong conceptual associations in Reddit polls. While we are aware of the existence of a variety of methods for filtering word-word relationships (like minimum spanning trees \cite{kenett2011global}), it is difficult to gain insights about the effects that these methods have over selecting conceptual links, so that we resorted to a simpler frequency-based pruning, instead. This led to the creation of 5 different co-occurrence networks, each one representing the structure of conceptual associations as expressed by authors in their notes/posts. A network visualisation, built with Gephi, is reported in Figure \ref{fig:fig_1} as an example. Notice how “I” and “you” are central in the semantic/syntactic structure extracted from suicide notes. Then, to further characterize the co-occurrence networks, we leverage closeness centrality. Closeness is a well-known centrality measure that identifies how close a node is to all other nodes in a network in terms of its average network distance \cite{siew2019cognitive}. In cognitive network science, network distance was shown to strongly correlate with semantic relatedness, so that concepts further apart on a semantic network were found to be less semantically related \cite{kenett2017semantic}. Hence, nodes with a higher closeness centrality are expected to be more semantically related to other connected concepts as mentioned in text, which was further confirmed by more recent studies \cite{stella2020text}. 
Taking inspiration from past approaches in cognitive network science \cite{stella2020lockdown}, we performed an emotional enrichment of co-occurrence networks. We endowed stems with valence and emotional attributes as coming from the Valence-Arousal-Dominance \cite{mohammad2018obtaining} and from the Emotion Lexicon \cite{mohammad2013crowdsourcing} datasets, respectively. Valence is a psychological metric estimating how pleasant/unpleasant concepts are perceived. 
Words were attributed "positive" (upper quartile), "negative" (lower quartile) or neutral (otherwise) labels, purely for visualisation purposes. 
Words were also attributed a list of emotions they elicited according to the cognitive data gathered by \cite{mohammad2013crowdsourcing}.
\subsection{Semantic frame theory and Latent Dirichlet analysis}
Co-occurrence networks are representations of the ways authors associate concepts in their own narratives \cite{carley1993coding, akimushkin2017text}. 
Considering a network neighbourhood for a word $W$ identifies all concepts associated to it in text, i.e. the semantic frame of associates providing meaning and content to $W$ itself \cite{fillmore2010frames}. 

Another way to reconstruct a semantic frame is to identify a topic, i.e. a class of highly co-occurring concepts in a set of documents. 
To implement this second characterisation of semantic frames we adopted Latent Dirichlet analysis, as implemented in the Gensim and pyLDAvis packages in Python. 
To fix the number $K$ of topics to search for we adopted information coming from network structure. 
We used the Louvain method \cite{blondel2008fast} to count how many communities C populated the semantic frame of “feel” in a given network.
We then used this number, |$C$|, as input for detecting $K$ = |$C$| topics via LDA, but only in documents mentioning at least once either “feel” or its different word forms (e.g. “feeling”). Since in LDA the same concept can be repeated across different topics, we selected as potential semantic frame the topic where “feel” occurred with the highest frequency.
Notice that this LDA analysis offers only a limited understanding of which words were associated with “feel”, so that we used it only to validate results coming from the network analysis.
In our further investigations of the emotions populating the semantic frame of “feel” across suicide letters and blog posts we focused over network neighbourhoods/semantic frames.
\subsection{Emotional profiling}
The emotions attributed to individual words \cite{mohammad2013crowdsourcing} were used to explore how individuals felt in their online narratives. 
Emotional profiling was applied to the semantic frame of “feel” across all reconstructed networks. 
For each semantic frame/network neighbourhood, we measured the fraction $r_{i}=m/N$ of $m$ words inspiring emotion $i$ in a neighbourhood with $N$ entries. 
The collection of all fractions $r_{i}$ constitutes the so-called emotional profile of a given semantic frame \cite{\cite{stella2020lockdown}}. 
Notice that word negations were considered in these counts: As in \cite{stella2020text}, antonyms of words linked with negations (like "not") were added to the count. 
Then, the emotional profile was matched against 1000 random counterparts, all built by sampling uniformly at random from the Emotion Lexicon the same number of words eliciting at least one emotion observed in the empirical profile. 
The empirical and random scores were used to compute $z$-scores indicating the strength of the observed emotion in a given semantic frame.
Z-scores were plotted as petals, in a visualisation reminiscent of Plutchik's wheel of emotions \cite{semeraro2021pyplutchik} and that we call emotional flower \cite{stella2020text}. 
Petals falling outside of a rejection region $z$ $<$ 1.96 (semi-transparent circles) are indicative of strong emotional intensities populating a given frame.
\section{Results}
We present our results starting from simple, unstructured frequency analyses in texts and semantic prominence rankings in cognitive networks. 
Finding evidence that authors of texts expressed their feelings in different ways across corpora, we move to more specific investigations of semantic frames for “feel”. 
We present key findings about different emotional profiles of “feel” between suicide notes and Reddit forums. We conclude by validating our results via a combination of community analysis and LDA.
\label{sec:results}
\subsection{Prominent concepts across suicide notes and clinical Reddit posts}

Table \ref{tab:tab1} reports the top-20 most semantically prominent concepts as measured by closeness centrality and frequency (see Section \ref{sec:methods}) across all the corpora. 
Results highlight that the first-person pronoun is the most central concept across all the considered networks, either according to frequency (of mentions) or closeness (i.e. semantic prominence). In suicide notes, “I love you” is the set of most prominent/frequent words: People who committed suicide expressed prominently their love to their dear ones. In the clinical subreddits, key concepts are relative to distress and feelings (e.g. “anxiety” being more frequent than “you” in r/Anxiety, “feel” being prominent/frequent across all clinical boards). 
This indicates a difference between suicide letters, addressing expressions of love, and subreddits, reporting personal feelings and experiences. 
Interestingly, this difference supports previous studies underlining strong self-attentional focus in narratives produced by clinically anxious individuals \cite{yoo2019semantic, teixeira2021revealing}. 
\begin{table}[h]
\centering
\begin{tabular}{@{}|c|c|c|c|c|c|c|c|c|c|@{}}
\toprule
\multicolumn{2}{|c|}{\textit{\textbf{Suicide Notes}}} &
  \multicolumn{2}{c|}{\textit{\textbf{Anxiety}}} &
  \multicolumn{2}{c|}{\textit{\textbf{Depression}}} &
  \multicolumn{2}{c|}{\textit{\textbf{Schizophrenia}}} &
  \multicolumn{2}{c|}{\textit{\textbf{DIY}}} \\ \midrule
\textit{Freq.} &
  \textit{Clos.} &
  \textit{Freq.} &
  \textit{Clos.} &
  \textit{Freq.} &
  \textit{Clos.} &
  \textit{Freq.} &
  \textit{Clos.} &
  \textit{Freq.} &
  \textit{Clos.} \\ \midrule
\textit{I} &
  \textit{I} &
  \textit{I} &
  \textit{I} &
  \textit{I} &
  \textit{I} &
  \textit{I} &
  \textit{I} &
  \textit{I} &
  \textit{I} \\ \midrule
\textit{you} &
  \textit{you} &
  \textit{anxiety} &
  \textit{anxiety} &
  \textit{you} &
  \textit{you} &
  \textit{he} &
  \textit{he} &
  \textit{wall} &
  \textit{wall} \\ \midrule
\textit{love} &
  \textit{love} &
  \textit{go} &
  \textit{go} &
  \textit{she} &
  \textit{she} &
  \textit{you} &
  \textit{you} &
  \textit{paint} &
  \textit{want} \\ \midrule
\textit{go} &
  \textit{go} &
  {\color[HTML]{FF0000} \textit{feel}} &
  {\color[HTML]{FF0000} \textit{feel}} &
  \textit{go} &
  \textit{go} &
  \textit{they} &
  \textit{they} &
  \textit{need} &
  \textit{paint} \\ \midrule
\textit{he} &
  \textit{he} &
  \textit{you} &
  \textit{you} &
  {\color[HTML]{FF0000} \textit{feel}} &
  {\color[HTML]{FF0000} \textit{feel}} &
  \textit{go} &
  \textit{go} &
  \textit{look} &
  \textit{need} \\ \midrule
\textit{will} &
  \textit{will} &
  \textit{time} &
  \textit{time} &
  \textit{time} &
  \textit{time} &
  \textit{she} &
  \textit{she} &
  \textit{use} &
  \textit{look} \\ \midrule
\textit{life} &
  \textit{life} &
  \textit{make} &
  \textit{make} &
  \textit{life} &
  \textit{life} &
  {\color[HTML]{FF0000} \textit{feel}} &
  {\color[HTML]{FF0000} \textit{feel}} &
  \textit{go} &
  \textit{use} \\ \midrule
\textit{please} &
  \textit{please} &
  \textit{she} &
  \textit{they} &
  \textit{depress} &
  \textit{depress} &
  \textit{think} &
  \textit{think} &
  \textit{new} &
  \textit{go} \\ \midrule
\textit{take} &
  \textit{take} &
  \textit{they} &
  \textit{she} &
  \textit{they} &
  \textit{they} &
  \textit{people} &
  \textit{people} &
  \textit{way} &
  \textit{new} \\ \midrule
\textit{way} &
  \textit{way} &
  \textit{think} &
  \textit{think} &
  \textit{friend} &
  \textit{friend} &
  \textit{make} &
  \textit{make} &
  \textit{no} &
  \textit{way} \\ \midrule
\textit{know} &
  \textit{no} &
  \textit{work} &
  \textit{work} &
  \textit{myself} &
  \textit{myself} &
  \textit{time} &
  \textit{time} &
  \textit{make} &
  \textit{no} \\ \midrule
\textit{want} &
  \textit{one} &
  \textit{really} &
  \textit{really} &
  \textit{make} &
  \textit{make} &
  \textit{up} &
  \textit{thing} &
  \textit{side} &
  \textit{make} \\ \midrule
\textit{make} &
  \textit{thing} &
  \textit{take} &
  \textit{take} &
  \textit{people} &
  \textit{people} &
  \textit{know} &
  \textit{know} &
  \textit{water} &
  \textit{side} \\ \midrule
\textit{help} &
  \textit{know} &
  \textit{he} &
  \textit{thing} &
  \textit{even} &
  \textit{even} &
  \textit{take} &
  \textit{take} &
  \textit{want} &
  \textit{aer} \\ \midrule
\textit{tell} &
  \textit{want} &
  \textit{people} &
  \textit{he} &
  \textit{want} &
  \textit{want} &
  \textit{really} &
  \textit{one} &
  \textit{work} &
  \textit{water} \\ \bottomrule
\end{tabular}%
\caption{Top 15 concepts ranked in the order of decreasing frequency in corpora and closeness centrality in networks. The concept “feel” is highlighted in red.}
\label{tab:tab1}
\end{table}
Bi-gram analysis confirms this difference between love-focused suicide letters and self-focused online posts (see Supplementary Table \ref{tab:tab4}). 
In addition, suicide notes are found to frequently mention bi-grams expressing hope and regret (e.g. “I - hope” or “I - sorry”, which misses the stopword “am” because of network construction), which are missing in subreddit texts. The latter narratives focus more on expressing individual needs and mental distress (e.g. “I - depress”, “I - feel” and “feel - like”). 

Overall, the above indicates that the texts analysed here crucially reported how individuals expressed their own feelings, perceptions and states, thus motivating further analysis.
\subsection{Cognitive networks quantify different feelings across suicide notes and clinical Reddit posts}
\begin{figure*}[h]
\centering
  {\includegraphics[scale=0.6]{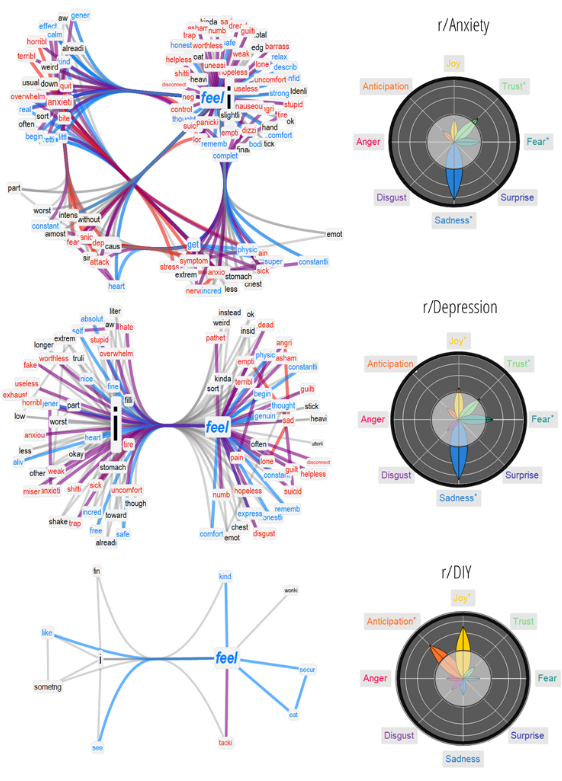}}
    \caption{Cognitive networks (left) and emotional flowers (right) for the semantic frames of “feel” across r/Anxiety (top), r/Depression (middle) and r/DIY (bottom). Links in cognitive networks indicate concept co-occurrence. Colours indicate valence, i.e. positive (cyan), negative (red) or neutral (black) valence. Links between positive/negative concepts are in cyan/red. Purple links connect concepts of contrasting valence. Flowers are made of petals representing z-scores of emotional counts (see Methods). Petals falling outside the semi-transparent rejection region (z < 1.96) indicate stronger than expected concentrations of concepts eliciting a given emotion.}
    \label{fig:fig_2}
\end{figure*}

\begin{figure*}[h]
\centering
  {\includegraphics[scale=0.6]{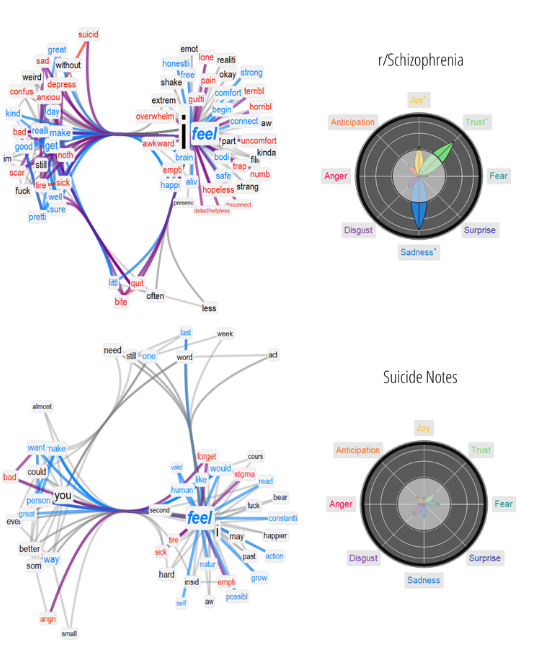}}
    \caption{Cognitive networks (left) and emotional flowers (right) for the semantic frames of “feel” across r/Schizophrenia (top) and suicide notes (bottom). Links in cognitive networks indicate concept co-occurrence. Colours indicate valence, i.e. positive (cyan), negative (red) or neutral (black) valence. Links between positive/negative concepts are in cyan/red. Purple links connect concepts of contrasting valence. Flowers are made of petals representing z-scores of emotional counts (see Methods). Petals falling outside the semi-transparent rejection region (z < 1.96) indicate stronger than expected concentrations of concepts eliciting a given emotion.}
    \label{fig:fig_3}
\end{figure*}

Figures \ref{fig:fig_2} and \ref{fig:fig_3} report the semantic and emotional content associated with “feel” in suicide letters and Reddit posts.
All figures report the semantic network community of “feel” (left), i.e. focusing on concepts most tightly associated with “feel”,  and the emotional flower relative to all associations framing “feel” (right). 
These results are based on our cognitive network of co-occurrences/emotional labels, capturing how text authors associated ideas and concepts in their narratives.

In r/Anxiety, Figure \ref{fig:fig_2} (top), authors use “feel” to express an overwhelmingly negative jargon, including several negative emotional states like “loneliness”, “guilt” and “helplessness”. 
Semantic content relative to “panic”, “hopelessness”, “stress”, concepts like being “worthless” and “overwhelmed” and mentions of body parts like “stomach” and “hand”, all strongly indicate mentions of panic attacks, which are frequently experienced by individuals suffering from anxiety disorders \cite{allen2010cognitive}. 
These concepts elicit strong levels of sadness, as reported in the emotional flower. However, this emotion co-exists with trust, i.e. a feeling of confidence and security in an idea \cite{dunn2005feeling}. 
Trust spawns from positive jargon linked with “feel”, like “calm”, “begin”, “kind” and “relax”, which indicate positive language mixing with negative ideas in subreddit posts. 
The co-existence of positive and negative concepts leads to the emotional polarisation between sadness and trust observed in the semantic frame of “feel” for users of r/Anxiety. 
Similar semantic content is observed for online users in the r/Depression board (Figure \ref{fig:fig_2}, middle), where the emotional polarisation is stronger, contrasting stronger-than-expected levels of sadness and fear against joy and trust. 
Fear spawns from jargon indicating uselessness and emptiness, e.g “fake”, “pathetic”, “dead”, and including mentions to other disorders like “anxiety” and “suicide”. 
These communicative patterns indicate the presence of a strong overlap in feelings between depression and other clinical conditions, as supported by clinical comorbidity studies \cite{batinic2017comorbidity}. 
Despite this commonality, users in r/Depression express their feelings with stronger fear and joy than users in r/Anxiety, underlining an important difference between these distress signals that is only recently being remarked by psychologists through network psychometrics \cite{o2021importance}.

In the r/DIY board (Figure \ref{fig:fig_2}, bottom), “feel” is more peripheral than in other boards: it has a lower degree and a more sparse semantic frame. “Feel” in r/DIY is framed with joy and anticipation, e.g. cheerful planning. 
This is expected from a board where users report non-clinical stories about home repairing and it indicates that our methods are not biased to always detect negative emotional states. 

Whereas feelings expressed in r/Depression and r/Anxiety feature sadness as being prevalent, r/Schizophrenia is more deeply polarised, featuring equally strong levels of sadness and trust simultaneously (see Figure \ref{fig:fig_3}, top). 
Schizophrenia is a severe mental illness where thinking and perceptions are strongly distorted, with symptoms like hallucinations, paranoia and disarticulated speech in addition to comorbidities with anxiety and depression \cite{mccutcheon2020schizophrenia}. 
This clinical connotation is validated by the narratives expressed by online users, which mention anxiety, depression and hopelessness. Interestingly, these negative concepts are mixed with positive/trusted ones, like “happy”, “alive”, “comfort”, “strong” and “good”. 
Notably, the semantic community of “feel” in r/Schizophrenia includes links with “connect”, “detached” and “disconnected”, which relate to the well-documented sense of isolation that hallucinations and paranoia can create in patients affected by schizophrenia \cite{green2018social}.

Suicide letters frame “feel” in a different way compared to all the above clinical subreddits (see Figure \ref{fig:fig_3}, bottom). 
No strong emotional signal is found in the semantic frame of “feel”, which features only a few negative associations (with “empty”, “tired” and “sick”) overwhelmed by generally positive jargon (e.g. “want”, “make”, “person”). 
This lack of emotional content indicates that suicide notes are substantially different from clinical subreddits. 
Our results indicate that whereas online forums showcase emotional signals and semantic content relative to personal diaries, where people report how they feel in relationship with their conditions, suicide notes are not informative about how individuals directly feel.
In fact, no expected distress pattern relative to suicide ideation, like anxiety or depression \cite{batinic2017comorbidity}, is found in narratives from suicide letters. 
These last notes frame “feel” in general terms, mentioning mostly people and actions, as expected in reports of events.

\subsection{Results of Topic Analysis}
To further support the above results about emotional polarisation and relationships of feelings with clinical conditions, we examined our textual corpora through a Latent Dirichlet Analysis (see Section \ref{sec:methods}). 
Table \ref{tab:tab2} reports words in the topic where “feel” was the most frequent, ranking them according to their frequency. The r/DIY board was discarded from the analysis as it showcased distinct patterns from other boards.

All clinical boards and suicide notes express feelings as related with the passage of time (e.g. “day”, “hour”, “time”).
Whereas clinical boards consistently mention social environments (e.g. “work”, “friend”, “people”), suicide notes prominently express feelings of love that are missing in clinical boards.
This confirms that suicide letters are more personal and less formal narratives, targeting loved ones rather than general online audiences. 
The co-occurrence of “feel”, “want”, “help” and “tell” in clinical subreddits indicate a willingness for authors to seek aid and comfort, which further confirms the semantic/emotional signals of distress and trust found in the previous section. 
As indicated by narrative psychology, narratives can help people understand and confront their issues \cite{murray2003narrative}. 
Our network and LDA analyses provide strong evidence that this help is explicitly mentioned in subreddits when individuals report their well being or support each other.

Suicide notes mention requests for help to a lesser extent than clinical subreddits, and rather focus feelings towards being “sorry” for having to “leave” and depart from “love”(d) ones (cf. Table \ref{tab:tab2}). 
This distinctive trait shifts emotional
connotations from expressions of feelings to final salutations, providing a reason for the absence of strong emotional signals from the network analysis. 
This also indicates that suicide notes happen at a later stage than mere diaries, when individuals have already processed their mental distress and do not want to express it anymore in their final words bidding farewell to loved ones.
\begin{table}[h]
\centering
\begin{tabular}{@{}|l|l|l|l|l|@{}}
\toprule
\textit{\textbf{Rank}} & \textit{\textbf{Suicide Notes}} & \textit{\textbf{r/Anxiety}} & \textit{\textbf{r/Depression}} & \textit{\textbf{r/Schizophrenia}} \\ \midrule
\textit{1}  & \textit{love}   & \textit{go}      & \textit{know}      & \textit{want}   \\ \midrule
\textit{2}  & \textit{go}     & \textit{know}    & \textit{want}      & \textit{know}   \\ \midrule
\textit{3}  & \textit{know}   & \textit{time}    & \textit{go}        & \textit{time}   \\ \midrule
\textit{4}  & \textit{want}   & \textit{think}   & \textit{depressed} & \textit{year}   \\ \midrule
\textit{5}  & \textit{please} & \textit{want}    & \textit{think}     & \textit{day}    \\ \midrule
\textit{6}  & \textit{life}   & \textit{really}  & \textit{time}      & \textit{really} \\ \midrule
\textit{7}  & \textit{take}   & \textit{thing}   & \textit{really}    & \textit{take}   \\ \midrule
\textit{8}  & \textit{live}   & \textit{year}    & \textit{life}      & \textit{try}    \\ \midrule
\textit{9}  & \textit{time}   & \textit{say}     & \textit{day}       & \textit{work}   \\ \midrule
\textit{10} & \textit{leave}  & \textit{anxiety} & \textit{year}      & \textit{life}   \\ \midrule
\textit{11} & \textit{way}    & \textit{life}    & \textit{try}       & \textit{even}   \\ \midrule
\textit{12} & \textit{money}  & \textit{try}     & \textit{work}      & \textit{tell}   \\ \midrule
\textit{13} & \textit{god}    & \textit{need}    & \textit{help}      & \textit{help}   \\ \midrule
\textit{14} & \textit{last}   & \textit{friend}  & \textit{people}    & \textit{friend} \\ \midrule
\textit{15} & \textit{sorry}  & \textit{tell}    & \textit{never}     & \textit{need}   \\ \bottomrule
\end{tabular}%
\caption{Top 15 concepts in the topic including “feel” across text corpora, as obtained from Latent Dirichlet Allocation. Stopwords were removed to focus on semantic content.}
\label{tab:tab2}
\end{table}

\section{Discussion}
In the present work, we used a combination of methods from cognitive network science and NLP to study how people express their feelings through mental health online boards (i.e., r/anxiety, r/depression, r/schizophrenia, and r/do-it-your-own) and 142 genuine suicide notes. 
In this final Section, we focus our attention on the emotional profiles derived from the word “feel” in these two different scenarios. 
Indeed, the rich emotional structure conveyed by this word reveals non-trivial, heterogeneous associations within the mental health communities but a different, more neutral profile in the suicide notes.

Focusing on the Reddit boards firstly, we observe that “feel” is not related to a unique, mono-dimensional emotion, or even to a primary dyad \cite{semeraro2021pyplutchik}: its semantic frame is populated by pairs of unexpected, often contrasting emotions, as in the polarised “sadness/trust” example from r/schizophrenia or, to a smaller extent, the “sadness/joy” pair from r/depression. 
It is also interesting to mention that only a “homogeneous” emotional dyad appears in r/DIY, i.e., the “anticipation/joy” dyad, indicating how mental health communities are more complex than other thematic subreddits where people ask questions to solve problems on their own. 
Such a dyad indeed is a marker of “optimism” \cite{semeraro2021pyplutchik}, an attitude that may fit well with people communicating their desire to do things by themselves. 
Conversely, the heterogeneous emotional profiles of “feel” in the clinical boards highlight how interactions on digital platforms are not all the same \cite{pavalanathan2015identity}, and a variety of users could be present within them for different reasons and needings. 
For instance, there can be both people asking for support and people giving that support, where the former ones amplify the “negative” petal of “feel”’s frame, and the latter ones represent the opposite emotional flower. 
An addendum for this, and also a parallel interpretation, can account for the presence of several users at different stages of a mental condition \cite{o2021importance}. 
Hence, users who need to express their feelings or ask for support can listen and perhaps learn from the previous experiences of other users. 
It is clear that only further analyses could validate such interpretations through additional data mining.       
Moving to suicide notes analysis, we find that people tend to express their feelings in a quite different way with respect to Reddit boards. 
No strong emotional concepts emerge in the semantic frame of “feel” while studying the network of suicide notes, neither positives nor negatives. 
The fact that people tend to express feelings in an emotionless way could suggest that suicide letters are not meant to report how individuals who are going to end their lives really feel.  
Indeed, as emerging from previous analyses of the semantic frames of “life” and “love” \cite{teixeira2021revealing}, people who are going to commit suicide are more prone to express their gratitude and love to their families and friends instead of explicitly mention how they feel. 

Accordingly, our analysis provides quantitative evidence that if we want to predict suicide ideation through expressions of feelings, we should concentrate on earlier stages of mental distress, as they can be revealed more prominently from online interactions in mental health communities rather than from suicide notes. The “online diary” nature of subreddit boards, for instance, makes them helpful tools for the early-stage prediction of mental distress \cite{shen2017detecting, low2020natural, teixeira2021revealing}.
Indeed, preliminary studies already exploited Reddit boards for monitoring shifts from mental health disorders to suicidal ideation \cite{de2016discovering, tadesse2020detection, gaur2021characterization}. 
A list of distinct markers characterizing these shifts was suggested in this regard, involving heightened self-attentional focus, poor linguistic coherence, and strong manifestation of anxiety, impulsiveness, hopelessness, and loneliness \cite{de2016discovering}. 
Some of them can fit with the negative emotional petals found in each board, i.e., hopelessness and loneliness moods, while others might explain the heterogeneity of the semantic frames, e.g., the poor linguistic coherence, whereas such positive emotions come from the same individuals sharing also their emotional distress. 
However, to the best of our knowledge, the only work that leverages semantic networks to analyze mental health discourse on Reddit boards is the one by Yoo and colleagues \cite{yoo2019semantic}.
Like us, the authors define the semantic networks of three mental health-related subreddits and thus rely on topic modeling techniques to find discourse patterns. 
Even if they do not analyze specific semantic frames, the results obtained in r/depression are in line with what we have observed. 
Indeed, on the one hand, they found a prevalence of  “sadness” feeling all over the network but also a significant amount of joyful and positive expressions. 
On the other, as in our study, topic analysis suggests the particular attention given by mental health users to time-related words (e.g., “weeks”, “past”, “years”) as well as school and friends issues (e.g., “no friends”, “social skills”, “high school”).

Nevertheless, there are some limitations to our current research that also motivates future works.
Our analysis aggregates time and individual users, providing a collective quantification of semantic frames that can be opened in future research.
Considering a longitudinal approach of time-evolving social interactions (e.g., users remove their accounts, new friendships are made) and accounting for feelings expressed by individuals both represent interesting directions for future research. 
Although we plan to extend our work by including the “time” features as well as tracking individual emotional evolutions over time by considering individual semantic networks, such a fine-grained perspective could lead to ethical and privacy issues as to preserve users’ anonymity, especially in online mental health studies.
Regarding the current work, the networks that we build aggregate multiple groups from different perspectives, so we overcome the anonymity issue by making it impossible to identify individual users. Then, future research calls for identifying valid approaches aiming at combining the fine-grained analyses we need for validating our interpretation and the issues related to user anonymity.  
\label{sec:discussion}

\section*{Acknowledgments}
This work is supported by the European Union – Horizon 2020 Program under the scheme "INFRAIA-01-2018-2019 – Integrating Activities for Advanced Communities", Grant Agreement n.871042, "SoBigData++: European Integrated Infrastructure for Social Mining and Big Data Analytics" (http://www.sobigdata.eu). \footnote{Last Accessed: Oct, 28, 2021}

\bibliographystyle{unsrt}  
\bibliography{references}  

\newpage

\section*{Supplementary Information}

\renewcommand{\thetable}{S\arabic{table}}
\begin{table}[h]
\centering
\resizebox{\textwidth}{!}{%
\begin{tabular}{@{}|c|c|c|c|@{}}
\toprule
\multicolumn{1}{|l|}{} &
  \textit{\textbf{\begin{tabular}[c]{@{}c@{}}Mean Local  \\ Clustering\end{tabular}}} &
  \textit{\textbf{Assortativity}} &
  \textit{\textbf{\begin{tabular}[c]{@{}c@{}}No. of topics/  communities\\ (Louvain)\end{tabular}}} \\ \midrule
\textit{\textbf{Suicide Notes}} & 0.198 & -0.075 & 18 \\ \midrule
\textit{\textbf{Anxiety}}       & 0.426 & -0.209 & 21 \\ \midrule
\textit{\textbf{Depression}}    & 0.432 & -0.213 & 14 \\ \midrule
\textit{\textbf{Schizophrenia}} & 0.401 & -0.209 & 22 \\ \midrule
\textit{\textbf{DIY}}           & 0.12  & -0.108 & 18 \\ \bottomrule
\end{tabular}%
}
\caption{ Network statistics as defined in Siew et al. (2019) for the different co-occurrence networks analysed in the main text.}
\end{table}

\begin{table}[h]
\centering
\resizebox{\textwidth}{!}{
\begin{tabular}{@{}|
>{\columncolor[HTML]{FFFFFF}}l |
>{\columncolor[HTML]{FFFFFF}}l |
>{\columncolor[HTML]{FFFFFF}}l |
>{\columncolor[HTML]{FFFFFF}}l |
>{\columncolor[HTML]{FFFFFF}}l |@{}}
\toprule
\textit{\textbf{Suicide Notes}} & \textit{\textbf{Anxiety}} & \textit{\textbf{Depression}} & \textit{\textbf{Schizophrenia}} & \textit{\textbf{DIY}} \\ \midrule
\textit{(love, you)} & \textit{(I, know)}    & \textit{(I, know)}    & \textit{(I, know)}    & \textit{(I, need)}  \\ \midrule
\textit{(I, love)}   & \textit{(feel, I)}    & \textit{(feel, I)}    & \textit{(I, think)}   & \textit{(I, want)}  \\ \midrule
\textit{(I, know)}   & \textit{(I, think)}   & \textit{(I, want)}    & \textit{(feel, I)}    & \textit{(I, use)}   \\ \midrule
\textit{(I, you)}    & \textit{(get, I)}     & \textit{(I, think)}   & \textit{(I, want)}    & \textit{(I, try)}   \\ \midrule
\textit{(I, sorry)}  & \textit{(I, like)}    & \textit{(feel, like)} & \textit{(get, I)}     & \textit{(I, look)}  \\ \midrule
\textit{(hope, I)}   & \textit{(feel, like)} & \textit{(I, like)}    & \textit{(go, I)}      & \textit{(go, I)}    \\ \midrule
\textit{(I, want)}   & \textit{(go, I)}      & \textit{(get, I)}     & \textit{(I, like)}    & \textit{(find, I)}  \\ \midrule
\textit{(go, I)}     & \textit{(I, want)}    & \textit{(go, I)}      & \textit{(feel, like)} & \textit{(I, would)} \\ \midrule
\textit{(know, you)} & \textit{(anxiety, I)} & \textit{(I, time)}    & \textit{(I, say)}     & \textit{(get, I)}   \\ \midrule
\textit{(I, think)}  & \textit{(I, time)}    & \textit{(I, really)}  & \textit{(I, time)}    & \textit{(I, know)}  \\ \bottomrule
\end{tabular}%
}
\caption{Most frequent bi-grams after stopword removal and stemming in texts.}
\label{tab:tab4}
\end{table}

\end{document}